\documentclass[10pt,twocolumn,letterpaper]{article}

\usepackage{cvpr}      % To produce the REVIEW version
\definecolor{cvprblue}{rgb}{0.21,0.49,0.74}
\usepackage[pagebackref,breaklinks,colorlinks,allcolors=cvprblue]{hyperref}
\usepackage{booktabs}
\usepackage{multirow}
\usepackage{amssymb}
\usepackage{url}
\usepackage{CJKutf8}
\def\confName{CVPR}
\def\confYear{2026}

\title{When AVSR Meets Video Conferencing: Dataset, Degradation, and the Hidden Mechanism Behind Performance Collapse}

\author{
    Yihuan Hunag$^{1,2}$\quad 
    Jun Xue$^{2}$\quad 
    Liu Jiajun$^{2}$\quad 
    Daixian Li$^{2}$\quad 
    Tong Zhang$^{2}$\quad \\[1.5mm]
    Zhuolin Yi$^{2}$\quad 
    Yanzhen Ren$^{1,2}$\thanks{Corresponding author.}\quad 
    Kai Li$^{3}$ \\[1.5mm]
    $^1$Key Laboratory of Aerospace Information Security and Trusted Computing, Ministry of Education\\
    $^2$School of Cyber Science and Engineering, Wuhan University\\
    $^3$Tsinghua University \\
    {\tt\small \{yihuanhuang, junxue, renyz\}@whu.edu.cn, tsinghua.kaili@gmail.com}
}

\begin{document}
\maketitle
\begin{abstract}
Audio-Visual Speech Recognition (AVSR) has achieved remarkable progress in offline conditions, yet its robustness in real-world video conferencing (VC) remains largely unexplored. This paper presents the first systematic evaluation of state-of-the-art AVSR models across mainstream VC platforms, revealing severe performance degradation caused by transmission distortions and spontaneous human hyper-expression. To address this gap, we construct \textbf{MLD-VC}, the first multimodal dataset tailored for VC, comprising 31 speakers, 22.79 hours of audio-visual data, and explicit use of the Lombard effect to enhance human hyper-expression. Through comprehensive analysis, we find that speech enhancement algorithms are the primary source of distribution shift, which alters the first and second formants of audio. Interestingly, we find that the distribution shift induced by the Lombard effect closely resembles that introduced by speech enhancement, which explains why models trained on Lombard data exhibit greater robustness in VC. Fine-tuning AVSR models on MLD-VC mitigates this issue, achieving an average 17.5\% reduction in CER across several VC platforms. Our findings and dataset provide a foundation for developing more robust and generalizable AVSR systems in real-world video conferencing. MLD-VC is available at \href{https://huggingface.co/datasets/nccm2p2/MLD-VC}{https://huggingface.co/datasets/nccm2p2/MLD-VC}.
\end{abstract}

\section{Introduction}
\label{sec:intro}
Audio-Visual Speech Recognition (AVSR) \cite{auto-avsr-1,auto-avsr-2,avsr_llm_1,robust_avsr_1} integrates audio and visual modalities to effectively overcome the performance limitations of single-modality Automatic Speech Recognition (ASR) \cite{asr_1,asr_2,asr_3,asr_4} under noisy or degraded conditions. It has become an important research direction for robust speech understanding. With the advancement of deep learning, state-of-the-art (SOTA) AVSR models have achieved remarkable performance on existing datasets. Since the outbreak of the COVID-19 pandemic, video conferencing (VC) platforms such as Zoom, Lark, Tencent Meeting, and DingTalk have become the primary means of remote communication. Consequently, AVSR has shown increasing demand in meeting transcription and accessibility support applications. 

\begin{figure}[t]
  \centering
   \includegraphics[width=1\linewidth]{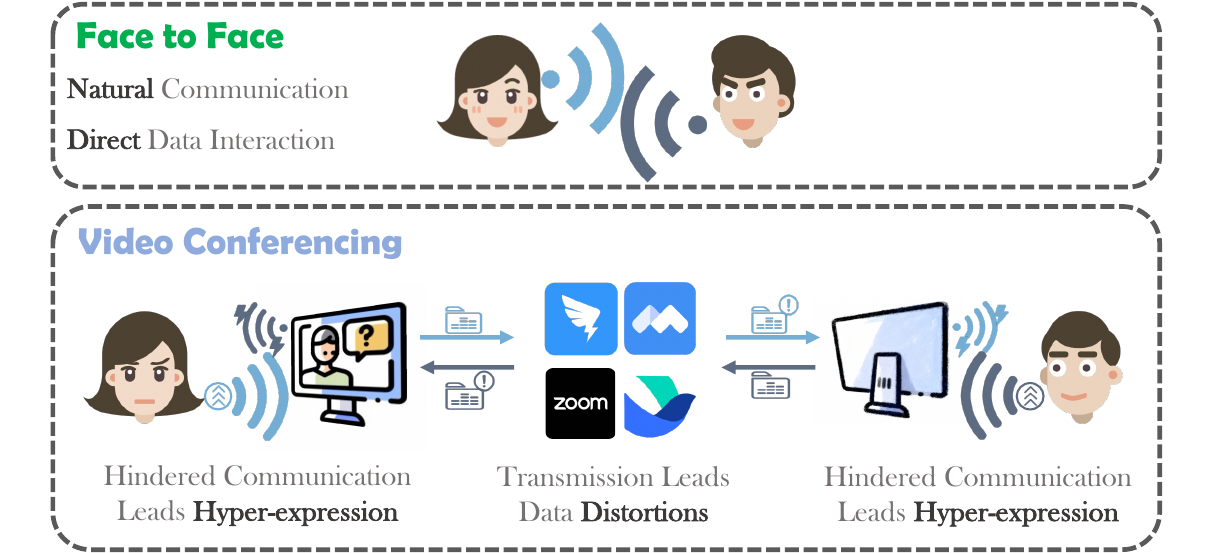}
   \caption{Compared to the face-to-face scenario, we identify two key factors affecting AVSR in video conferencing: transmission distortions in online and hyper-expression in a hindered communication environment.}
   \label{application}
\end{figure}

Most studies on robust AVSR \cite{robust_avsr_1,robust_avsr_2,robust_avsr_3,robust_avsr_4,robust_avsr_5} focus on noisy conditions or modality loss. However, our systematic evaluation reveals a critical issue: AVSR models experience severe performance degradation in VC, where the Word Error Rate (WER) and Character Error Rate (CER) increase from 0.93\%/0.56\% to 33.09\%/33.01\%. This collapse stems from the lack of datasets that reflect real-world VC conditions, as existing robustness studies are primarily conducted in offline or simulated environments, which fail to capture the complexity of real-world usage. As shown in Fig.~\ref{application}, communication in VC occurs through a camera and a display. This setting prompts participants to implicitly assume that the ongoing interaction is more constrained than face-to-face communication, leading them to adjust their communicative behaviors spontaneously. In addition, both audio and visual signals undergo compression, noise suppression, and speech enhancement during transmission, which introduces distortions to the data.

In this paper, we present the first systematic evaluation of mainstream AVSR models in VC, construct the first multimodal dataset covering multiple platforms, and reveal the issue of acoustic feature drift through a detailed analysis of the proposed dataset. By evaluating AVSR models across various VC platforms and datasets, we find consistent performance degradation across platforms, modalities, and languages, highlighting the universality of this problem. Based on the evaluation results and analysis, we identify two key factors responsible for the degradation of AVSR performance in VC: transmission distortions and spontaneous human hyper-expression \cite{hyper_1,hyper_theory}. To bridge the research gap in VC, we construct the first multimodal dataset for video conferencing (MLD-VC). MLD-VC comprises 31 speakers and 22 hours of recordings covering four mainstream VC platforms, with audio, video, and lip landmark data. Moreover, MLD-VC explicitly incorporates the Lombard effect to enhance and capture the manifestation of hyper-expression in VC. Experimental results show that fine-tuning AVSR models on MLD-VC reduces the average CER by 17.5\% in VC.

We summarize the main contributions of this paper as follows.
\begin{enumerate}
    \item [1)] \textbf{Conducting the First Systematic Evaluation in VC.} We conduct the first systematic evaluation of AVSR models in VC, revealing the widespread nature of performance degradation. We further identify two key contributing factors to this degradation: transmission distortions and spontaneous human hyper-expression.

    \item [2)] \textbf{Constructing the First Multimodal VC Dataset for AVSR.} We construct the first multimodal dataset captured directly through multiple real VC platforms (MLD-VC), which not only considers the real-world VC scenario but also incorporates the Lombard effect to enhance the hyper-expression. MLD-VC comprises 4 mainstream VC platforms, 31 speakers, and 22.79 hours of audio-visual recordings. 

    \item [3)] \textbf{Revealing the Hidden Mechanism Behind Performance Collapse.} We reveal that the fundamental cause of AVSR performance degradation in VC is feature distribution shift. We further demonstrate that speech enhancement algorithms are the primary factor driving the shift in audio distributions. In addition, we find that landmark-level features in the visual modality remain stable in VC, providing new insights for current AVSR visual encoders that rely on the unstable image-level representations.

    \item [4)] \textbf{Improving AVSR Performance in VC.} After fine-tuning the AVSR model on MLD-VC, we achieve an average reduction of 17.5\% in CER across several VC platforms. The ablation results show that the two key contributing factors are indispensable for improving model performance.
\end{enumerate}

\section{Related Work}
\subsection{Audio-Visual Speech Recognition}
AVSR \cite{avsr_1,avsr_2,avsr_3,avsr_4,avsr_5,avsr_6,avsr_7,auto-avsr-1,auto-avsr-2,avsr_llm_1,avsr_llm_2,avsr_llm_3,avsr_llm_4} integrates visual and acoustic features to enhance recognition accuracy under challenging and interference-prone conditions. Most existing studies on AVSR primarily focus on model robustness under noisy conditions or modality degradation \cite{robust_avsr_1,robust_avsr_2,robust_avsr_3,robust_avsr_4,robust_avsr_5}. 

Since the outbreak of the COVID-19 pandemic, AVSR systems have been widely adopted in VC scenarios. In such scenarios, besides background noise and modality loss, audio-visual streams are further affected by compression distortion and the spontaneous human hyper-expression \cite{hyper_1}. These factors introduce challenges that extend beyond traditional noise-robust settings, yet remain largely overlooked by current research. Most existing works still rely on pre-collected and tightly synchronized datasets, without accounting for the compression artifacts and hyper-expression behaviors commonly observed in VC \cite{hyper_1}. Consequently, the generalization of existing robust AVSR methods to real-world VC scenarios remains severely limited. However, there is currently a lack of datasets for AVSR robustness in VC scenarios.

\subsection{Human Hyper-expression}
Hyper-expression refers to a speaker’s compensatory behavior when communication is hindered \cite{hyper_1,hyper_theory, lombard_2,lombard_3,lombard_4}. In such cases, the speaker tends to increase vocal intensity, exaggerate facial expressions, and use more gestures to enhance intelligibility. Lindblom et al. \cite{hyper_theory} proposed the hyper/hypo theory, which explains the underlying mechanism of hyper-expression. According to this theory, speakers continuously evaluate the communicative environment to adjust their articulations dynamically in response to contextual demands. A typical form of hyper-expression is the Lombard effect, which occurs when speakers attempt to communicate in noisy environments \cite{lombard_2,lombard_3,lombard_4,lombard_avsr_1,lombard_avsr_2}. 
% In addition, the work of Trujillo et al. \cite{lombard_vision} demonstrated that hyper-expression can also be triggered by visual disturbances, suggesting that both visual and acoustic interferences can jointly elicit such behavior.

Russell et al. \cite{hyper_1} further showed that hyper-expression is prevalent in VC scenarios. By analyzing real-world Zoom meeting recordings, they found that participants exhibited hyper-expression, which is similar to the Lombard effect. The specific characteristics are more frequent pauses, longer vowel durations, and higher fundamental frequency values. These findings suggest that the performance degradation of AVSR models in VC scenarios is not solely attributed to the online communication setting, but also stems from the spontaneous hyper-expression of speakers.

\section{AVSR Performance in Video Conferencing}
\begin{table*}[htbp]
  \centering
  \caption{Performance of AVSR models on video conferencing platforms. “Offline” refers to the original dataset. “*” indicates that the corresponding metric is not applicable or not reported. “$\Delta$” refers to the absolute difference between VC platforms and Offline. Bold indicates the least change for each platform.}
  \label{avsr_results}
  \scalebox{0.7}{
  \setlength{\tabcolsep}{5pt}
  \begin{tabular}{ccccccccc}
    \toprule
    \textbf{Model} & \textbf{Dataset} & \textbf{Modal} & \textbf{Language} & \textbf{Platform} & \textbf{WER(\%)↓} & \textbf{$\Delta$ WER(\%)↓} & \textbf{CER(\%)↓} & \textbf{$\Delta$ CER(\%)↓} \\
    \midrule
    \multirow{8}{*}{mWhisper-Flamingo \cite{mWhisper-Flamingo}} 
      & \multirow{8}{*}{LRS3 \cite{lrs3}} 
      & \multirow{4}{*}{A} & \multirow{4}{*}{En} 
      & Offline & 0.68 & * & * & * \\
      &  &  &  & Zoom  & 10.38 & 9.70 & * & * \\
      &  &  &  & Lark  & 19.55 & 18.87 & * & * \\
      &  &  &  & Tencent Meeting  & 11.89 & 11.21 & * & * \\
      \cmidrule(lr){3-9}
      &  & \multirow{4}{*}{AV} & \multirow{4}{*}{En}  & Offline & 0.73 & * & * & * \\
      &  &  &  & Zoom  & 9.22 & 8.49 & * & * \\
      &  &  &  & Lark  & 18.53 & 17.80 & * & * \\
      &  &  &  & Tencent Meeting  & 10.97 & 10.24 & * & * \\
    \midrule
    \multirow{8}{*}{LiPS-AVSR \cite{LiPS-AVSR}} 
      & \multirow{8}{*}{Chinese-Lips \cite{LiPS-AVSR}} 
      & \multirow{4}{*}{A} & \multirow{4}{*}{Zh} 
      & Offline & * & * & 4.37 & * \\
      &  &  &  & Zoom  & * & * & 8.67 & \textbf{4.30} \\
      &  &  &  & Lark  & * & * & 14.83 & 10.46 \\
      &  &  &  & Tencent Meeting  & * & * & 9.72 & 5.35 \\
      \cmidrule(lr){3-9}
      &  & \multirow{4}{*}{AV} & \multirow{4}{*}{Zh} & Offline & * & * & 3.87 & * \\
      &  &  &  & Zoom  & * & * & 18.53 & 14.66 \\
      &  &  &  & Lark  & * & * & 10.97 & 7.10 \\
      &  &  &  & Tencent Meeting  & * & * & 9.22 & 5.35 \\
    \midrule
    \multirow{12}{*}{Auto-AVSR \cite{auto-avsr-2}} 
      & \multirow{8}{*}{LRS3 \cite{lrs3}} 
      & \multirow{4}{*}{AV} & \multirow{4}{*}{En} 
      & Offline & 0.93 & * & 0.56 & * \\
      &  &  &  & Zoom  & 33.09 & 32.16 & 33.01 & 32.45 \\
      &  &  &  & Lark  & 31.72 & 30.79 & 30.71 & 30.15 \\
      &  &  &  & Tencent Meeting  & 23.79 & 22.86 & 22.65 & 22.09 \\
      \cmidrule(lr){3-9}
      & 
      & \multirow{4}{*}{V} & \multirow{4}{*}{En} 
      & Offline & 19.08 & * & 13.68 & * \\
      &  &  &  & Zoom  & 90.26 & 71.18 & 74.32 & 60.64 \\
      &  &  &  & Lark  & 92.08 & 73.00 & 74.51 & 60.83 \\
      &  &  &  & Tencent Meeting  & 56.14 & 37.06 & 43.88 & 30.20 \\
      \cmidrule(lr){2-9}
      & \multirow{4}{*}{Lombard-Grid \cite{lombard_grid}} 
      & \multirow{4}{*}{AV} & \multirow{4}{*}{En} 
      & Offline & 4.93 & * & 3.06 & * \\
      &  &  &  & Zoom  & 12.36 & \textbf{7.43} & 9.93 & 6.87 \\
      &  &  &  & Lark  & 8.94 & \textbf{4.01} & 7.79 & \textbf{4.73} \\
      &  &  &  & Tencent Meeting  & 7.12 & \textbf{2.19} & 4.48 & \textbf{1.42} \\
    \bottomrule
  \end{tabular}
  }
\end{table*}

With the spread of the COVID-19 pandemic, VC has become commonplace, whereas it was previously a relatively niche practice. AVSR models are often deployed on VC platforms to transcribe the meeting content. Existing AVSR models are typically trained on clean datasets, and artificial noise is later added during post-processing to improve model robustness. However, real-world VC scenarios involve multiple complex factors, including codec compression, transmission delays, and speech enhancement effects. This section aims to systematically evaluate the performance of existing SOTA models under VC conditions.

\subsection{Setup}\label{setup}
To comprehensively evaluate the performance of current AVSR models under VC conditions, we employed three representative VC platforms and evaluated three SOTA AVSR models across multiple datasets and languages. 

\subsubsection{Dataset}
Considering the linguistic differences, we selected both English and Chinese multimodal datasets, specifically Lombard-Grid \cite{lombard_grid}, LRS3 \cite{lrs3}, and Chinese-Lips \cite{LiPS-AVSR}. 

% Lombard-Grid is a multimodal corpus covering both frontal and profile views, consisting of 5400 video samples from 54 speakers, including 2700 Lombard and 2700 plain reference utterances, with each video following the standardized sentence structure of the original “Grid” corpus. The LRS3 dataset contains over 400 hours of face tracks extracted from TED and TEDx talks, along with their corresponding textual transcriptions. Chinese-Lips is a Chinese multimodal audio-visual dataset comprising 100 hours of facial video clips paired with their corresponding sentence transcriptions.

\subsubsection{Baselines}
To evaluate the performance of AVSR models in VC, we selected three SOTA models, specifically Auto-AVSR \cite{auto-avsr-2}, mWhisper-Flamingo \cite{mWhisper-Flamingo}, and LiPS-AVSR \cite{LiPS-AVSR}. 
% Auto-AVSR is trained on the Lombard-Grid and LRS3 datasets. mWhisper-Flamingo is built upon the LRS3 dataset. LiPS-AVSR is trained on the Chinese-Lips dataset. 

\subsection{Metrics}
We adopt Word Error Rate (WER) and Character Error Rate (CER) as metrics. Lower WER and CER values indicate better performance.

\subsubsection{Video Conferencing Platform}
With the widespread adoption of VC, various applications have rapidly emerged. We selected several commonly used platforms, including Zoom, Lark, and Tencent Meeting. 

\subsection{Evaluation Method}
To ensure a fair comparison of baselines across different platforms, we followed the open-source configurations of each baseline, successfully reproducing their performance with results closely matching those reported in the original papers. Furthermore, we transmitted the test sets of each corresponding dataset through the VC platforms to simulate the VC conditions. We provide a detailed description of the transmission process in Appendix 8.1. We then computed the WER and CER on the transmitted test sets to evaluate the performance of each baseline. 

\subsection{Quantitative Results} \label{previous_results}
\subsubsection{Performance Analysis}
% We present the quantitative results in Tab. \ref{avsr-results} and visualize the WER/CER increment caused by video conferencing transmission as heatmaps in Fig. \ref{increase}. 

We present the quantitative results in Tab. \ref{avsr_results}. The results reveal that all models suffer significant performance degradation under VC conditions. This degradation remains consistent across different languages and modalities, confirming the destructive impact of VC on AVSR models. mWhisper-Flamingo exhibits a significant performance decline when deployed in VC compared to its offline baseline. On the LRS3 dataset, the WER of mWhisper-Flamingo in audio-only modality increases by an average of 20.5 times compared to the offline setting. Although the audio-visual modality variant slightly mitigates the degradation (WER = 9.22\% on Zoom vs. 10.38\% in audio-only), the overall accuracy remains significantly lower than that of the offline condition. 

For LiPS-AVSR, the degradation pattern is consistent. The offline CER of 4.37\% (audio-only) and 3.87\% (audio-visual) nearly doubles or triples under conferencing conditions. Notably, the audio-visual configuration suffers substantial instability, reaching up to 18.53\% on Zoom. This suggests that visual cues may be unreliable when video compression or transmission delay occurs.

The Auto-AVSR model exhibits the most pronounced performance degradation among the three evaluated systems. On the LRS3 dataset, the WER of Auto-AVSR in audio-visual modality increases from 0.93\% to 33.09\%. In contrast, the audio-visual evaluation on the Lombard-Grid dataset shows relatively minor degradation, indicating that Auto-AVSR trained with Lombard data is more resilient to VC distortions. Through the comparison between the Lombard-Grid and LRS3 datasets, it can be observed that the model trained on Lombard data is inherently more stable under conferencing conditions.

\subsubsection{Modal Analysis}
Different modality combinations exhibit markedly distinct levels of tolerance to VC conditions, which directly determine the robustness of AVSR models. Among them, the visual-only modality is the most vulnerable, as all datasets show severe performance degradation under VC scenarios. For instance, in the Lark of the LRS3 dataset, the Auto-AVSR model achieves a WER of 92.08\%, while in the Zoom it reaches 90.26\%. These results indicate that the visual-only modality is more sensitive to compression and delay during transmission, making it difficult for a vision-only system to resist such distortions. 

Audio-only modality exhibits moderate robustness, with the WER/CER increases of the mWhisper-Flamingo and LiPS-AVSR models under VC conditions being lower than those of the audio-visual modality. For instance, in the LRS3 dataset under Zoom transmission, the mWhisper-Flamingo model achieves a WER of 10.38\% in audio-only modality, higher than 9.22\% in audio-visual modality. 

Audio-visual modality exhibits the strongest robustness. In most cases, the audio-visual modality of baseline models effectively mitigates interference in VC scenarios through cross-modal information complementarity. For the Auto-AVSR model on the LRS3 dataset, the AV fusion modality achieves a maximum WER of 33.09\% in VC scenarios, which is substantially lower than the 92.08\% observed in the visual-only modality. The audio-visual modality of the mWhisper-Flamingo model achieves lower WER than the audio-only modality across all three platforms.

\subsubsection{Dataset Analysis}
The inherent characteristics of different datasets (such as video quality, speaker diversity, and data distribution) directly affect the extent of performance degradation after transmission. Among them, the Lombard-Grid dataset shows the most substantial resistance to interference. For the Auto-AVSR model on this dataset, the WER under the audio-visual modality remains as low as 12.36\% even after transmission via Zoom, which is substantially lower than the 33.09\% WER observed for the same modality on the LRS3 dataset. In addition, the model trained on Lombard-Grid exhibits the smallest increases in WER and CER across nearly all settings. The primary reason lies in the fact that half of the data in Lombard-Grid exhibits the Lombard effect. Previous studies \cite{lombard_avsr_1, lombard_avsr_2} have confirmed that training AVSR models on data containing the Lombard effect can simultaneously improve recognition accuracy and robustness.

\subsection{Summary of Findings}\label{findings}
We evaluated several AVSR models across mainstream VC platforms. We found that the performance degradation of AVSR models is a pervasive issue, regardless of model architecture, language, modality, or VC platform. Interestingly, we observed that models trained with Lombard data exhibit strong robustness in VC.

\section{Multimodal Dataset for Video Conferencing}
The previous section revealed that AVSR models degrade notably in VC. However, existing datasets fail to capture the unique characteristics of such scenarios. In this section, we identify two key factors of VC scenarios and construct the first multimodal VC dataset for AVSR (MLD-VC), which explicitly incorporates these two key factors.
\subsection{Key Factors of Video Conferencing}
According to the previous analysis, we identify two key factors of VC scenarios, as shown in Fig.~\ref{application}. \textbf{(K1) Transmission distortions.} Compared to offline scenarios, audio-visual signals in VC are subject to complex interferences, including codec compression, noise suppression, and speech enhancement processing. These external interferences degrade the clarity and intelligibility of both audio and visual streams. \textbf{(K2) Spontaneous human hyper-expression.} Beyond distortion, human behavior in VC differs from offline scenarios. Psychoacoustic and auditory studies \cite{hyper_1,hyper_theory,lombard_3} have demonstrated that speakers, under conditions of communicative obstruction, spontaneously enhance their vocal expressions and facial articulations to ensure effective information transmission. This phenomenon, termed “Hyper-expression”, manifests as a slower speech rate, expanded vowel articulation space, and more pronounced lip and head movements. Previous studies \cite{hyper_1} have confirmed the widespread presence of such hyper-expression behaviors in VC. Since the Lombard effect is a typical form of hyper-expression \cite{hyper_theory}, this finding explains why the models trained with Lombard data in Section \ref{findings} exhibit stronger robustness in VC.

Under the combined influence of \textbf{K1} and \textbf{K2}, the data distribution in VC deviates significantly from that of “clean” in offline conditions. For AVSR models trained on offline data, these two key factors directly cause a severe degradation in model performance.

\subsection{Dataset Construction}
\subsubsection{Participant information and consent}
We recruited a total of 31 volunteers to participate in the data collection process, all of whom were university students, including 15 males and 16 females. Each participant was fully informed about the research objectives, data recording procedures, and the specific types of personal information retained (e.g., age and gender). All participants were required to sign an informed consent form that clearly stated the intended use of the collected data. After the data collection, participants were asked to review all recorded audio-visual materials for verification. The final released dataset will undergo a rigorous anonymization process to ensure participant privacy.
\subsubsection{Core of Dataset}\label{innovation}
To better highlight the characteristics of VC, we systematically incorporate \textbf{K1} and \textbf{K2} in the proposed dataset. To faithfully reproduce transmission distortions under VC conditions, we select several widely used platforms, including Tencent Meeting, Lark, DingTalk, and Zoom. Previous studies \cite{hyper_1} have shown that hyper-expression behaviors in VC are highly analogous to the Lombard effect. The Lombard effect \cite{lombard_avsr_1, lombard_avsr_2} is induced by environmental noise, and its intensity varies in relation to the level of noise exposure. Therefore, we introduce controlled noise stimuli to elicit the Lombard effect in speakers, thereby enhancing the representativeness of hyper-expression phenomena in VC.

By incorporating the two key factors of VC, our constructed dataset more accurately aligns with the signal properties and human interaction patterns of VC. This dataset can make a new contribution to investigating the degradation mechanisms and enhancing the robustness of AVSR models under VC conditions.

\subsubsection{Corpus Design}
We designed both an English and a Chinese corpus. The sentence construction was inspired by the Lombard-Grid dataset (English) \cite{lombard_grid} and the DB-MMLC dataset (Chinese) \cite{lombard_2}, both of which adopt the Grid-style grammar. Each English sentence in our dataset consists of six words. For example, “bin blue at A 2 please” follows the structure:
\textless Command: bin, lay, place, set\textgreater \textless Color: blue, green, red, white\textgreater \textless Preposition: at, by, in, with\textgreater  \textless Letter: A–Z (excluding W)\textgreater \textless Digit: 0–9\textgreater \textless Adverb: again, now, please, soon\textgreater. Among them, three words (color, letter, and digit) are regarded as keywords, while the remaining ones are considered fillers. The Chinese version of the Grid-style corpus used in our dataset is provided in Appendix 8.2.1.

To introduce the Lombard effect, we followed previous studies and designed four background noise conditions, including Plain (no noise), 40 dB, 60 dB, and 80 dB. All participants were informed that they were taking part in a video conference. They were required to wear headphones, sit in front of a display and camera, and read aloud the sentences from the corpus. The display prompted participants to start reading, while the corresponding background noise was simultaneously played through the headphones. Participants were required to complete the reading task under each of the noise conditions. Each participant was required to read 30 sentences (20 Chinese sentences and 10 English sentences) under each of the four background noise conditions. 

\subsubsection{Recording} \label{recording}
To simulate real-world usage scenarios, all VC platforms were configured with their default settings. The audio and video input devices consisted of a UGREEN CM564 microphone and a UGREEN CM831-65381 camera. Both the input and receiving audio–video streams were recorded using OBS Studio. The recording parameters of OBS Studio were set as follows: stereo audio with a 48 kHz sampling rate encoded using the AAC codec, and video with a resolution of 1920 × 1080 at 25 FPS. The recordings from the input side were treated as offline data, while those from the receiving side were regarded as VC data.

\subsubsection{Post Processing}
To improve the quality of the dataset, we manually inspected the recorded raw data and discarded any “corrupted samples” caused by unexpected recording issues. To ensure anonymization, we followed the preprocessing configuration of Auto-AVSR \cite{auto-avsr-2} and cropped each video to retain only the lip region. To eliminate the influence of system volume during recording, we applied loudness normalization to the audio. To ensure compatibility with various AVSR model inputs, we used FFMPEG to convert all audio into single-channel WAV files with a sampling rate of 16 kHz.

\subsection{Dataset Distribution}
% Table generated by Excel2LaTeX from sheet 'Sheet1'
\begin{table}[t]
  \centering
    \caption{
    Comparison of the proposed MLD-VC with existing multimodal Lombard and video conferencing datasets. “VC Num” refers to the number of video conferencing platforms included. “*” indicates that the dataset was collected offline.
  }
  \resizebox{\linewidth}{!}{
    \begin{tabular}{cccccc}
    \toprule
    \textbf{Dataset} & \textbf{Speakers} & \textbf{Duration(h)} & \textbf{Language} & \textbf{VC Num} & \textbf{Year} \\
    \midrule
    Lombard-Grid & 54    & 3.81     & En    & * & 2018 \\
    RoomReader & \textbf{118 }  & 8     & En    & 1  & 2022 \\
    \textbf{MLD-VC (Ours)}  & 31    &\textbf{ 22.79}    & \textbf{En \& Zh} & \textbf{4} & \textbf{2025} \\
    \bottomrule
    \end{tabular}%
  \label{dataset_compare}%
  }
\end{table}%

Tab. \ref{dataset_compare} compares our dataset with existing multimodal Lombard datasets and video conferencing datasets. Our dataset substantially extends the total duration and the number of platforms compared with existing datasets. Compared to RoomReader, our dataset is 2.8 times longer in total duration and encompasses three additional VC platforms. Compared with Lombard-Grid, our dataset is more than 6 times longer in total duration and extends to the online scenario. Additionally, our dataset supports multiple languages and various VC platforms. We present the duration of each subset across different VC platforms in Appendix 8.2.3.

\section{Revealing the Cause of AVSR Degradation}
In the previous results presented in Section \ref{previous_results}, we observed a significant performance degradation of AVSR models in VC. This phenomenon suggests that both the audio-visual features undergo distributional shifts under VC conditions, resulting in impaired performance. To further investigate this issue, we conducted a systematic analysis of audio-visual data in the proposed MLD-VC dataset. We compared the characteristics of audio and visual modalities between offline and online settings, revealing the direct manifestations of AVSR performance degradation. In addition, we dissected the transmission and processing pipeline in VC systems to identify the critical stages responsible for modality distribution shifts, thereby uncovering the underlying causes of AVSR performance deterioration. 

\subsection{Analysis of Modality Difference}
To analyze the modality discrepancies across different scenarios, we visualize the probability density distributions of representative features. For the audio modality, we use openSMILE \cite{eyben2010opensmile} to extract five acoustic features: the fundamental frequency (F0), the first formant (F1), the second formant (F2), loudness, and the ratio of the total energy in the 50-1kHz to the total energy in the 1kHz-5kHz (AlphaRatio). Note that the values of AlphaRatio are logarithmically scaled. 

For the visual modality, compression and network latency in online transmission inevitably cause quality degradation and even blurring. Traditional image quality measures, such as PSNR and SSIM, have limited value in this scenario because they cannot capture the information that AVSR models primarily rely on. The essential objective of AVSR is to recognize lip movements. Based on this task property, we use lip geometric motion as the primary target for analysis. Specifically, we use three indicators: lip width, lip height, and lip roundness. Lip roundness is defined as the ratio between height and width, and a value closer to 1 indicates a more rounded lip shape. All indicators are computed from facial landmark positions, and the detailed procedure is provided in Appendix 8.3.1.

\subsubsection{Difference between Offline and Online}
We present the probability density curves of speech features in Fig. \ref{audio_analysis}. In addition, Tab. \ref{peak_x_coordinates} lists the horizontal coordinates corresponding to the peak points of each curve. We find that the F0 remains nearly consistent across offline and online scenarios, indicating that the overall pitch level of speech is not substantially affected. In contrast, both F1 and F2 exhibit significant upward frequency shifts, with DingTalk showing the approximate 170 Hz increase. AlphaRatio in online settings is lower than in offline settings, indicating that high-frequency energy is enhanced in online recordings. Moreover, the loudness distribution shifts to the left in online scenarios, suggesting a slight attenuation in overall acoustic energy. These patterns are consistent across all VC platforms, indicating that VC platforms systematically alter the spectral structure of audio, thereby impairing the accurate modeling of features by AVSR models.

\subsubsection{Difference between Plain and Hyper-expression}
\begin{figure}[t]
  \centering
   \includegraphics[width=1\linewidth]{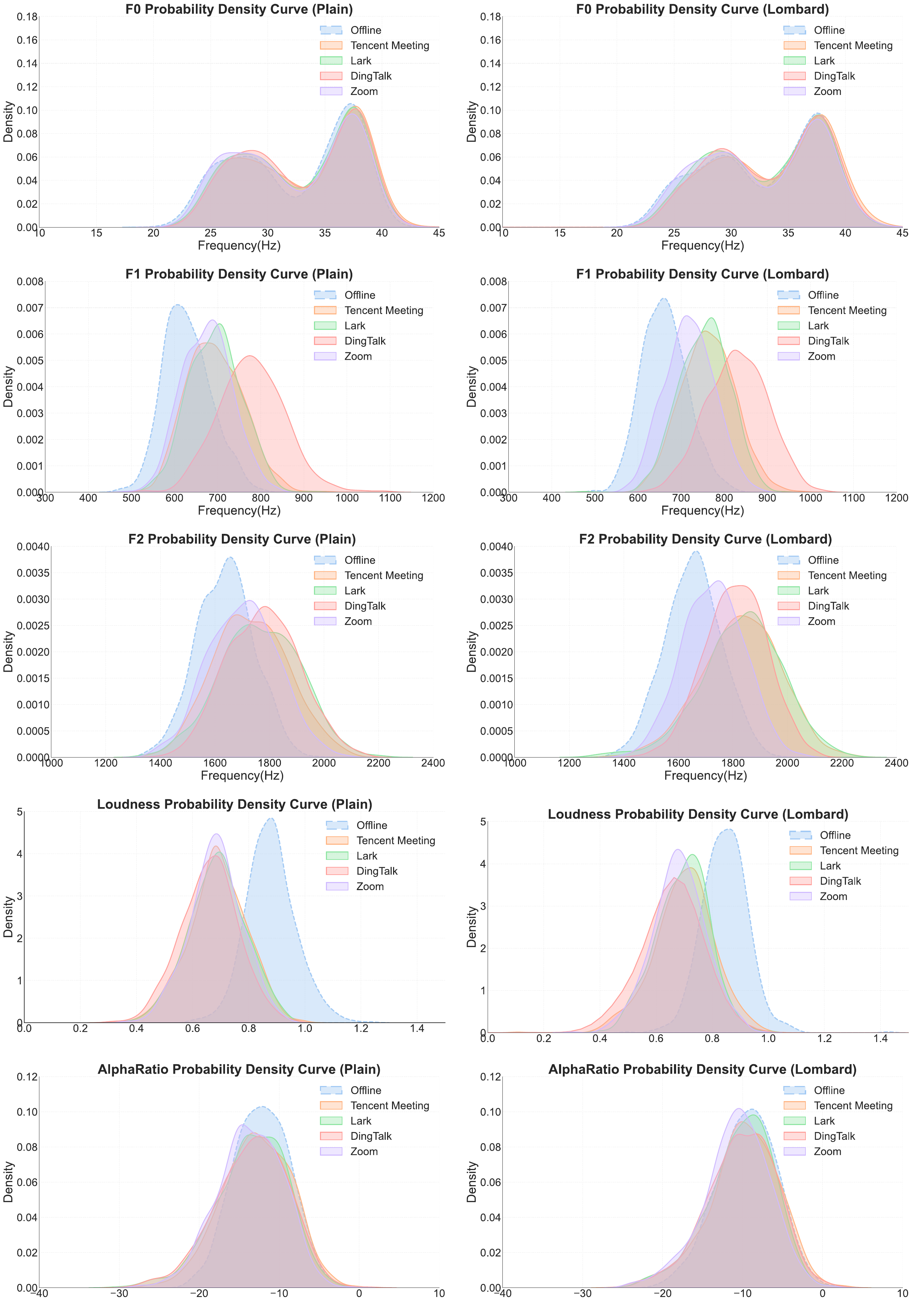}
\caption{
Probability density curves of five acoustic features (F0, F1, F2, Loudness, and AlphaRatio) across five subsets in the proposed MLD-VC under \textit{Plain} (left column) and \textit{Lombard} (right column).
Across features(column-wise), both \textit{Plain} and \textit{Lombard} show noticeable high-frequency shifts in F1 and F2 under video conferencing conditions.
Across conditions (row-wise), \textit{Lombard} also exhibits overall higher F1 and F2 than \textit{Plain}.
}
   \label{audio_analysis}
\end{figure}

We use the Lombard effect to enhance the hyper-expression in VC explicitly. By comparing the paired subplots (row-wise) in Fig. \ref{audio_analysis} with the corresponding sub-tables in Tab. \ref{peak_x_coordinates}, it can be observed that the F0 remains essentially unchanged under the Lombard condition. At the same time, the distributions of the F1 and F2 shift toward higher frequencies with reduced dispersion. We further observe that the variations of F1 and F2 between Plain and Lombard speech resemble those between offline and online recordings, indicating that the spectral structures induced by the Lombard effect are similar to those of video conferencing. This finding explains why the model trained with Lombard data in Section \ref{findings} shows stronger robustness in VC.

\subsubsection{Difference between Audio and Vision}
We present the visual feature analysis results in Appendix 8.3.2. The results show that the differences in visual features are minimal, indicating that the VC scenario exerts only a negligible influence on lip landmarks. However, this does not imply that the degradation in AVSR performance is unrelated to the visual modality. We observe that Auto-AVSR, mWhisper-Flamingo, and LiPS-AVSR employ pre-trained ResNet18 \cite{resnet18} and AVHuBERT \cite{shi2022avhubert} backbones to process visual inputs. Moreover, these models take lip images rather than lip landmarks as visual inputs. Due to codec compression, transmission delay, and other factors, lip images are subject to distortion, resulting in a distributional shift in the visual modality. In contrast, our analysis indicates that landmark-level features remain stable in VC, suggesting that future AVSR models could benefit from geometry-based visual encodings.

\begin{figure}[t]
  \centering
   \includegraphics[width=1\linewidth]{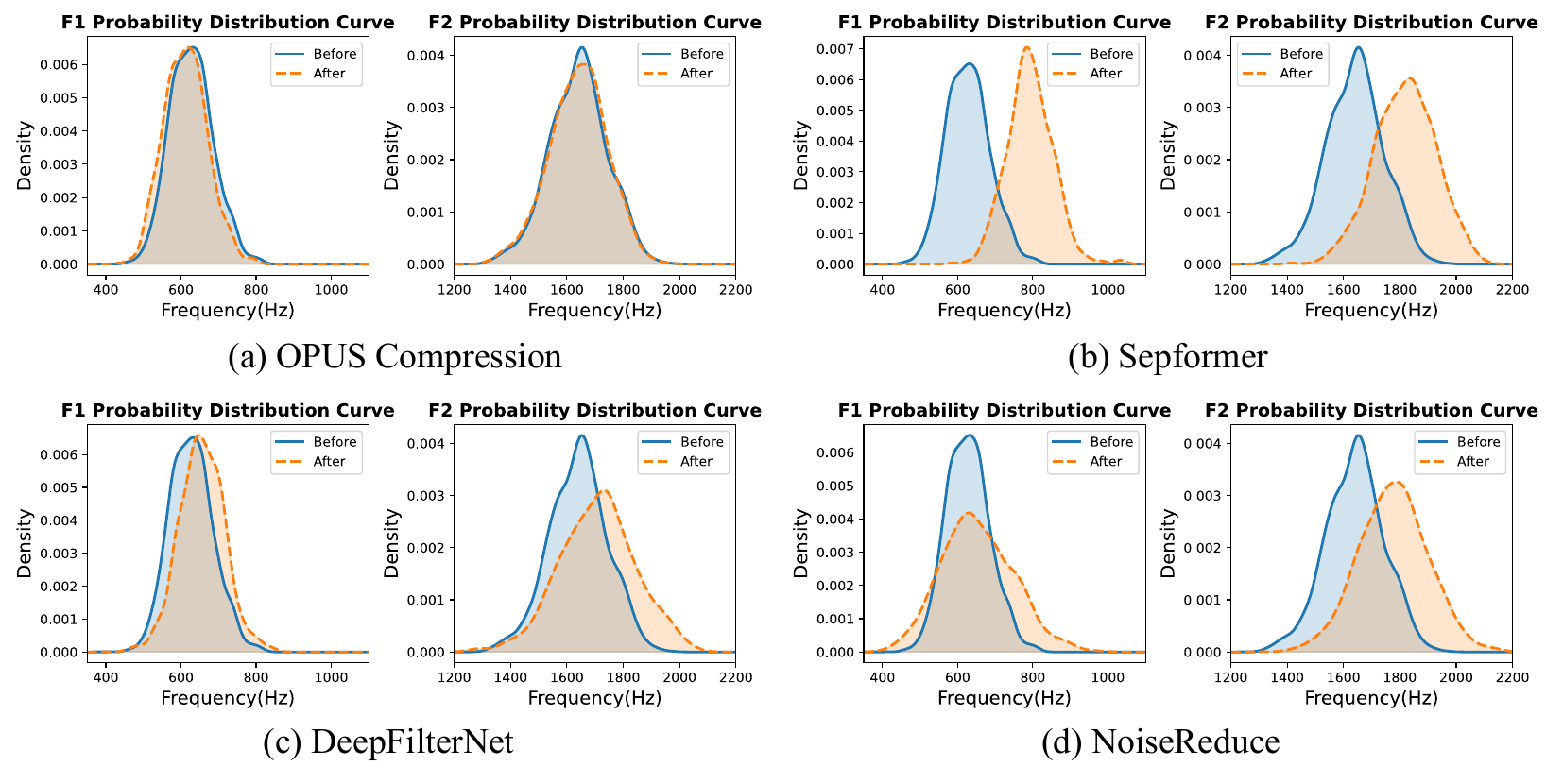}
   \caption{
Illustration of how video conferencing platforms affect speech formants.
We simulate platform processing by applying OPUS compression and three typical speech enhancement algorithms.
The four subplots show F1 and F2 distributions before and after processing.
Only the enhancement stages introduce noticeable spectral shifts and formant distortions, which explain the frequency bias observed in real video conferencing recordings.
}
\label{problem_location}
\end{figure}

\subsection{Identifying the Source of Distribution Drift}
\begin{table}[t]
\centering
\caption{Peak abscissas of probability density curves for different acoustic features.}
\label{peak_x_coordinates}
\resizebox{\linewidth}{!}{%
\begin{tabular}{cccccc}
\toprule
\multirow{2}{*}{\textbf{Acoustic Feature}} & \multicolumn{5}{c}{\textbf{Platform}} \\
\cmidrule(lr){2-6}
 & \textbf{Offline} & \textbf{Tencent Meeting} & \textbf{Lark} & \textbf{DingTalk} & \textbf{Zoom} \\
\midrule
F0 (\textit{Plain}) & 37.28  & 37.66  & 37.51  & 37.66  & 37.39  \\
F0 (\textit{Lombard})  & 37.58  & 37.85  & 37.62  & 37.64  & 37.55  \\
\midrule
F1 (\textit{Plain}) & 606.90  & 674.43  & 704.19  & 774.61  & 687.88  \\
F1 (\textit{Lombard})  & 660.06  & 756.11  & 770.17  & 825.44  & 712.73  \\
\midrule
F2 (\textit{Plain}) & 1655.66  & 1680.78  & 1727.37  & 1783.51  & 1727.45  \\
F2 (\textit{Lombard})  & 1665.07  & 1832.85  & 1863.88  & 1829.73  & 1746.69  \\
\midrule
Loudness (\textit{Plain}) & 0.88 & 0.68 & 0.69 & 0.68 & 0.68 \\
Loudness (\textit{Lombard})  & 0.86 & 0.72 & 0.73 & 0.67 & 0.68 \\
\midrule
AlphaRatio (\textit{Plain}) & -12.12 & -13.19 & -13.67 & -12.52 & -14.59 \\
AlphaRatio (\textit{Lombard})  & -8.81 & -10.07 & -8.67 & -8.28 & -10.47 \\
\bottomrule
\end{tabular}%
}
\end{table}

After analyzing Fig. \ref{audio_analysis}, we found that several acoustic features exhibited noticeable shifts under VC conditions. In particular, under the Plain condition, although hyper-expression occurred in the VC scenario, the shifts in F1 and F2 were significantly larger than those caused by hyper-expression. Therefore, we believe that the hyper-expression does not solely cause the pronounced deviations in F1 and F2 but is also likely influenced by the audio processing applied in the VC platforms.

To identify the underlying causes, we systematically deconstructed the speech processing pipeline in the VC scenario. Typically, the original speech undergoes codec compression and speech enhancement before being transmitted to the receiver. Since VC platforms operate as a black box, we can only approximate the process to locate the stages responsible for the acoustic feature shifts. We used the OPUS codec, which is widely adopted in VC platforms, such as Zoom, to simulate the codec compression stage. In addition, we employed Sepformer \cite{sepformer}, NoiseReduce \cite{noisereduce}, and DeepFilterNet \cite{schroeter2023deepfilternet3} as speech enhancement methods to model the enhancement stage in VC. The offline samples in MLD-VC were processed through codec compression and speech enhancement individually, and the resulting changes in their F1 and F2 were analyzed and compared.

Following the settings in Fig. \ref{audio_analysis}, we present the results in Fig. \ref{problem_location}. We observe that the codec compression stage has minimal impact on F1 and F2, with their frequency distributions remaining stable. In contrast, speech enhancement leads to an overall upward shift of F1 and F2. This phenomenon closely resembles the patterns observed in the real VC scenario illustrated in Fig. \ref{audio_analysis}. Therefore, we conclude that speech enhancement primarily causes the acoustic abnormalities in VC speech. While these algorithms improve the intelligibility, they also alter the spectral structure of speech. Consequently, these changes degrade the performance of AVSR. This finding reveals an essential cause of performance degradation in AVSR within the VC scenario from an acoustic perspective.

\section{Performance Evaluation on MLD-VC}
\subsection{Experiment Setup}
To further investigate the impact of MLD-VC for VC, we fine-tuned the AVSR model using the proposed MLD-VC dataset and evaluated its performance across different VC platforms. We followed the experimental setup described in Section \ref{setup} and utilized the LiPS-AVSR model \cite{LiPS-AVSR} for fine-tuning. The MLD-VC dataset was divided into training and testing sets. Additionally, ablation experiments were conducted to assess the contribution of online data and hyper-expression data to model performance. We present the implementation details in Appendix 8.4.
% \subsection{Implement Detials}

\subsection{Fine-tuning Results}
\begin{table}[t]
  \centering
  \caption{Results of fine-tuning with MLD-VC dataset on AVSR performance under video conferencing conditions. “*” indicates that the corresponding metric is not applicable or not reported.}
  \resizebox{\linewidth}{!}{
  \begin{tabular}{lcccc}
    \toprule
    \textbf{Test Dataset} & \textbf{Platform} & \textbf{Finetune} & \textbf{CER(\%)↓} & \textbf{Reduction(\%)} \\
    \midrule
    \multirow{6}{*}{Chinese-Lips \cite{LiPS-AVSR}} 
      & Tencent Meeting   & × & 10.97 & * \\
      & Tencent Meeting   & \checkmark & \textbf{9.65} & \textbf{12.0} \\
      & Lark & × & 18.53 & * \\
      & Lark & \checkmark & \textbf{13.64} & \textbf{26.4} \\
      & Zoom & × & 9.22 & * \\
      & Zoom & \checkmark & \textbf{7.93} & \textbf{14.0} \\
    \midrule
    \multirow{2}{*}{MLD-VC (Ours)} 
      & * & × & 42.37 & * \\
      & * & \checkmark & \textbf{13.91} & \textbf{67.2} \\
    \bottomrule
  \end{tabular}
  \label{tab:finetune}
  }
\end{table}

Tab.~\ref{tab:finetune} presents the fine-tuning results. We observed that models fine-tuned with MLD-VC achieved improved recognition accuracy across all VC platforms. Specifically, the fine-tuned models achieved an average relative reduction of 17.5\% in the CER across the three platforms. Furthermore, the CER on the MLD-VC test set decreased by 67.2\% after fine-tuning. This substantial improvement indicates that the fine-tuned models not only improve in-domain performance but also significantly enhance cross-platform generalization. These results highlight the effectiveness of MLD-VC for AVSR models in VC.

\subsection{Ablation}
\begin{table}[t] 
\footnotesize
  \centering
  \caption{Ablation study on the effects of online data and hyper-expression in MLD-VC fine-tuning. The results show that both online recording conditions and hyper-expression samples significantly contribute to lowering the CER.}
  \resizebox{\linewidth}{!}{
  \begin{tabular}{ccccc}
    \toprule
    \textbf{Online} & \textbf{Hyper-expression} & \textbf{Tencent Meeting} & \textbf{Lark} & \textbf{Zoom} \\
    \midrule
    \checkmark & \checkmark & \textbf{9.65} & \textbf{13.64} & \textbf{7.93} \\
    × & \checkmark & 10.15 & 15.52 & 10.53 \\
    \checkmark & × & 10.01 & 14.48 & 9.61 \\
    \bottomrule
  \end{tabular}
  \label{tab:ablation}
}
\end{table}

To better understand the contribution of different data components within the proposed MLD-VC dataset, we conduct an ablation study focusing on two key factors: (1) transmission distortions (online vs. offline) and (2) spontaneous human hyper-expression. 

The ablation results are summarized in Tab.~\ref{tab:ablation}. When online data were excluded, the average CER across VC platforms increased by 15.9\%. When hyper-expression data were excluded, the average CER increased by 10.5\%. These results indicate that both realistic recording conditions and spontaneous hyper-expressions are indispensable for improving the robustness of AVSR in VC scenarios. The combination of the two enables models to generalize more effectively to real-world VC conditions, validating the design rationale of the MLD-VC dataset discussed in Section \ref{innovation}.

\section{Conclusion}
We present the first systematic investigation of AVSR in real-world video conferencing. Our study reveals that transmission distortions and spontaneous human hyper-expression jointly lead to drastic performance degradation. Through detailed analysis, we identify speech enhancement algorithms as the primary source of distribution shift, which alters the F1/F2 formants. Moreover, we find that the distributional characteristics induced by the Lombard effect closely resemble those caused by speech enhancement, which explains why Lombard-trained models exhibit superior robustness in video conferencing. To bridge the gaps of lack of video conferencing data, we construct \textbf{MLD-VC}, the first multimodal dataset tailored for video conferencing, explicitly modeling hyper-expression under realistic online recording conditions. Fine-tuning AVSR models on MLD-VC achieves a 17.5\% average reduction in CER across platforms, and ablation studies confirm that both transmission distortions and hyper-expression data are crucial for improving the robustness of AVSR in video conferencing.

\section{Acknowledgement}
This work is supported in part by the National Natural Science Foundation of China (NO.62572358, NO.62172306, NO.62372334) and  DiDi Chuxing Group.

{
    \small
    \bibliographystyle{unsrt}
    \bibliography{main}

\begin{thebibliography}{10}

\bibitem{auto-avsr-1}
Pingchuan Ma, Alexandros Haliassos, Adriana Fernandez-Lopez, Honglie Chen, Stavros Petridis, and Maja Pantic.
\newblock Auto-avsr: Audio-visual speech recognition with automatic labels.
\newblock In {\em ICASSP 2023-2023 IEEE International Conference on Acoustics, Speech and Signal Processing (ICASSP)}, pages 1--5. IEEE, 2023.

\bibitem{auto-avsr-2}
Pingchuan Ma, Stavros Petridis, and Maja Pantic.
\newblock {Visual Speech Recognition for Multiple Languages in the Wild}.
\newblock {\em {Nature Machine Intelligence}}, 4:930--939, 2022.

\bibitem{avsr_llm_1}
Umberto Cappellazzo, Minsu Kim, Stavros Petridis, Daniele Falavigna, and Alessio Brutti.
\newblock Scaling and enhancing llm-based avsr: A sparse mixture of projectors approach.
\newblock {\em arXiv preprint arXiv:2505.14336}, 2025.

\bibitem{robust_avsr_1}
Joanna Hong, Minsu Kim, Jeongsoo Choi, and Yong~Man Ro.
\newblock Watch or listen: Robust audio-visual speech recognition with visual corruption modeling and reliability scoring.
\newblock In {\em Proceedings of the IEEE/CVF Conference on Computer Vision and Pattern Recognition}, pages 18783--18794, 2023.

\bibitem{asr_1}
Yufeng Yang, Ashutosh Pandey, and DeLiang Wang.
\newblock Towards decoupling frontend enhancement and backend recognition in monaural robust asr.
\newblock {\em Computer Speech \& Language}, 95:101821, 2026.

\bibitem{asr_2}
Yue Gu, Zhihao Du, Ying Shi, Shiliang Zhang, Qian Chen, and Jiqing Han.
\newblock Enhancing the robustness of contextual asr to varying biasing information volumes through purified semantic correlation joint modeling.
\newblock {\em IEEE Transactions on Audio, Speech and Language Processing}, 2025.

\bibitem{asr_3}
Rao Ma, Mengjie Qian, Mark Gales, and Kate Knill.
\newblock Asr error correction using large language models.
\newblock {\em IEEE Transactions on Audio, Speech and Language Processing}, 2025.

\bibitem{asr_4}
Hao Shi, Masato Mimura, and Tatsuya Kawahara.
\newblock Waveform-domain speech enhancement using spectrogram encoding for robust speech recognition.
\newblock {\em IEEE/ACM Transactions on Audio, Speech, and Language Processing}, 32:3049--3060, 2024.

\bibitem{robust_avsr_2}
Liangfa Wei, Jie Zhang, Junfeng Hou, and Lirong Dai.
\newblock Attentive fusion enhanced audio-visual encoding for transformer based robust speech recognition.
\newblock In {\em 2020 Asia-Pacific Signal and Information Processing Association Annual Summit and Conference (APSIPA ASC)}, pages 638--643. IEEE, 2020.

\bibitem{robust_avsr_3}
Jiahong Li, Chenda Li, Yifei Wu, and Yanmin Qian.
\newblock Robust audio-visual asr with unified cross-modal attention.
\newblock In {\em ICASSP 2023-2023 IEEE International Conference on Acoustics, Speech and Signal Processing (ICASSP)}, pages 1--5. IEEE, 2023.

\bibitem{robust_avsr_4}
Yusheng Dai, Hang Chen, Jun Du, Ruoyu Wang, Shihao Chen, Haotian Wang, and Chin-Hui Lee.
\newblock A study of dropout-induced modality bias on robustness to missing video frames for audio-visual speech recognition.
\newblock In {\em Proceedings of the IEEE/CVF Conference on Computer Vision and Pattern Recognition}, pages 27445--27455, 2024.

\bibitem{robust_avsr_5}
Maxime Burchi, Krishna~C Puvvada, Jagadeesh Balam, Boris Ginsburg, and Radu Timofte.
\newblock Multilingual audio-visual speech recognition with hybrid ctc/rnn-t fast conformer.
\newblock In {\em ICASSP 2024-2024 IEEE International Conference on Acoustics, Speech and Signal Processing (ICASSP)}, pages 10211--10215. IEEE, 2024.

\bibitem{hyper_1}
Sam~O’Connor Russell, Ayushi Pandey, and Naomi Harte.
\newblock Do we hyperarticulate on zoom?
\newblock In {\em Proc. CHiME 2023}, pages 77--81, 2023.

\bibitem{hyper_theory}
Bj{\"o}rn Lindblom.
\newblock Explaining phonetic variation: A sketch of the h\&h theory.
\newblock In {\em Speech production and speech modelling}, pages 403--439. Springer, 1990.

\bibitem{avsr_1}
Pingchuan Ma, Stavros Petridis, and Maja Pantic.
\newblock End-to-end audio-visual speech recognition with conformers.
\newblock In {\em ICASSP 2021-2021 IEEE International Conference on Acoustics, Speech and Signal Processing (ICASSP)}, pages 7613--7617. IEEE, 2021.

\bibitem{avsr_2}
George Sterpu, Christian Saam, and Naomi Harte.
\newblock Attention-based audio-visual fusion for robust automatic speech recognition.
\newblock In {\em Proceedings of the 20th ACM International conference on Multimodal Interaction}, pages 111--115, 2018.

\bibitem{avsr_3}
Yifei Wu, Chenda Li, Song Yang, Zhongqin Wu, and Yanmin Qian.
\newblock Audio-visual multi-talker speech recognition in a cocktail party.
\newblock In {\em Interspeech}, pages 3021--3025, 2021.

\bibitem{avsr_4}
Tao Li, Haodong Zhou, Jie Wang, Qingyang Hong, and Lin Li.
\newblock The xmu system for audio-visual diarization and recognition in misp challenge 2022.
\newblock In {\em ICASSP 2023-2023 IEEE International Conference on Acoustics, Speech and Signal Processing (ICASSP)}, pages 1--2. IEEE, 2023.

\bibitem{avsr_5}
Xinyu Wang, Haotian Jiang, Haolin Huang, Yu~Fang, Mengjie Xu, and Qian Wang.
\newblock Dcim-avsr: Efficient audio-visual speech recognition via dual conformer interaction module.
\newblock In {\em ICASSP 2025-2025 IEEE International Conference on Acoustics, Speech and Signal Processing (ICASSP)}, pages 1--5. IEEE, 2025.

\bibitem{avsr_6}
Fan Yu, Haoxu Wang, Ziyang Ma, and Shiliang Zhang.
\newblock Hourglass-avsr: Down-up sampling-based computational efficiency model for audio-visual speech recognition.
\newblock In {\em ICASSP 2024-2024 IEEE International Conference on Acoustics, Speech and Signal Processing (ICASSP)}, pages 7940--7944. IEEE, 2024.

\bibitem{avsr_7}
Fang Zhang, Yongxin Zhu, Xiangxiang Wang, Huang Chen, Xing Sun, and Linli Xu.
\newblock Visual hallucination elevates speech recognition.
\newblock In {\em Proceedings of the AAAI Conference on Artificial Intelligence}, volume~38, pages 19542--19550, 2024.

\bibitem{avsr_llm_2}
Umberto Cappellazzo, Minsu Kim, Honglie Chen, Pingchuan Ma, Stavros Petridis, Daniele Falavigna, Alessio Brutti, and Maja Pantic.
\newblock Large language models are strong audio-visual speech recognition learners.
\newblock In {\em ICASSP 2025-2025 IEEE International Conference on Acoustics, Speech and Signal Processing (ICASSP)}, pages 1--5. IEEE, 2025.

\bibitem{avsr_llm_3}
Jeong~Hun Yeo, Minsu Kim, Chae~Won Kim, Stavros Petridis, and Yong~Man Ro.
\newblock Zero-avsr: Zero-shot audio-visual speech recognition with llms by learning language-agnostic speech representations.
\newblock {\em arXiv preprint arXiv:2503.06273}, 2025.

\bibitem{avsr_llm_4}
Umberto Cappellazzo, Stavros Petridis, Maja Pantic, et~al.
\newblock Mitigating attention sinks and massive activations in audio-visual speech recognition with llms.
\newblock {\em arXiv preprint arXiv:2510.22603}, 2025.

\bibitem{lombard_2}
Hongcheng Zhu, Zongkun Sun, Yanzhen Ren, Kun He, Yongpeng Yan, Zixuan Wang, Wuyang Liu, Yuhong Yang, and Weiping Tu.
\newblock Lombard-vld: Voice liveness detection based on human auditory feedback.
\newblock In {\em 2025 IEEE Symposium on Security and Privacy (SP)}, pages 4303--4320. IEEE, 2025.

\bibitem{lombard_3}
Outi Tuomainen, Linda Taschenberger, Stuart Rosen, and Valerie Hazan.
\newblock Speech modifications in interactive speech: effects of age, sex and noise type.
\newblock {\em Philosophical Transactions of the Royal Society B}, 377(1841):20200398, 2022.

\bibitem{lombard_4}
James Trujillo, Asli {\"O}zy{\"u}rek, Judith Holler, and Linda Drijvers.
\newblock Speakers exhibit a multimodal lombard effect in noise.
\newblock {\em Scientific reports}, 11(1):16721, 2021.

\bibitem{lombard_avsr_1}
Ricard Marxer, Jon Barker, Najwa Alghamdi, and Steve Maddock.
\newblock The impact of the lombard effect on audio and visual speech recognition systems.
\newblock {\em Speech communication}, 100:58--68, 2018.

\bibitem{lombard_avsr_2}
Pingchuan Ma, Stavros Petridis, and Maja Pantic.
\newblock Investigating the lombard effect influence on end-to-end audio-visual speech recognition.
\newblock {\em arXiv preprint arXiv:1906.02112}, 2019.

\bibitem{mWhisper-Flamingo}
Andrew Rouditchenko, Samuel Thomas, Hilde Kuehne, Rogerio Feris, and James Glass.
\newblock mwhisper-flamingo for multilingual audio-visual noise-robust speech recognition.
\newblock {\em IEEE Signal Processing Letters}, 2025.

\bibitem{lrs3}
Triantafyllos Afouras, Joon~Son Chung, and Andrew Zisserman.
\newblock Lrs3-ted: a large-scale dataset for visual speech recognition.
\newblock {\em arXiv preprint arXiv:1809.00496}, 2018.

\bibitem{LiPS-AVSR}
Jinghua Zhao, Yuhang Jia, Shiyao Wang, Jiaming Zhou, Hui Wang, and Yong Qin.
\newblock Chinese-lips: A chinese audio-visual speech recognition dataset with lip-reading and presentation slides.
\newblock {\em arXiv preprint arXiv:2504.15066}, 2025.

\bibitem{lombard_grid}
Najwa Alghamdi, Steve Maddock, Ricard Marxer, Jon Barker, and Guy~J Brown.
\newblock A corpus of audio-visual lombard speech with frontal and profile views.
\newblock {\em The Journal of the Acoustical Society of America}, 143(6):EL523--EL529, 2018.

\bibitem{eyben2010opensmile}
Florian Eyben, Martin W{\"o}llmer, and Bj{\"o}rn Schuller.
\newblock Opensmile: the munich versatile and fast open-source audio feature extractor.
\newblock In {\em Proceedings of the 18th ACM international conference on Multimedia}, pages 1459--1462, 2010.

\bibitem{resnet18}
Kaiming He, Xiangyu Zhang, Shaoqing Ren, and Jian Sun.
\newblock Deep residual learning for image recognition.
\newblock In {\em Proceedings of the IEEE Conference on Computer Vision and Pattern Recognition (CVPR)}, June 2016.

\bibitem{shi2022avhubert}
Bowen Shi, Wei-Ning Hsu, Kushal Lakhotia, and Abdelrahman Mohamed.
\newblock Learning audio-visual speech representation by masked multimodal cluster prediction.
\newblock {\em arXiv preprint arXiv:2201.02184}, 2022.

\bibitem{sepformer}
Ui-Hyeop Shin, Sangyoun Lee, Taehan Kim, and Hyung-Min Park.
\newblock Separate and reconstruct: Asymmetric encoder-decoder for speech separation.
\newblock In {\em The Thirty-eighth Annual Conference on Neural Information Processing Systems}, 2024.

\bibitem{noisereduce}
Tim Sainburg, Marvin Thielk, and Timothy~Q Gentner.
\newblock Finding, visualizing, and quantifying latent structure across diverse animal vocal repertoires.
\newblock {\em PLoS computational biology}, 16(10):e1008228, 2020.

\bibitem{schroeter2023deepfilternet3}
Hendrik Schröter, Tobias Rosenkranz, Alberto~N. Escalante-B., and Andreas Maier.
\newblock {DeepFilterNet}: Perceptually motivated real-time speech enhancement.
\newblock In {\em INTERSPEECH}, 2023.

\end{thebibliography}
}
\clearpage
\setcounter{page}{1}
\maketitlesupplementary

\section{Appendix}
\label{Appendix}

\subsection{Detialed Transmission Process}
We transmit the test sets of LRS3 (1321 videos), Lombard-Grid (585 videos), and Chinese-Lips (3908 videos) through three video conferencing platforms, including Tencent Meeting, Lark, and Zoom. Specifically, we first concatenate all videos in each test set into a single long video. Between two consecutive videos, we insert a 0.4 second pure black and pure white clip as a segmentation flag. We then start a video conferencing session. On the sender side, we use the virtual camera of OBS as the camera input of the session. On the receiver side, we use OBS screen recording to capture the transmitted video. Note that our capture pipeline and parameter settings on the receiver side are kept identical to those described in Section \ref{recording}. For the transmitted videos, we cut them into individual clips that correspond to the original datasets according to the segmentation flags. We then re-encode these clips using the codec and resolution settings of the original datasets to ensure consistency.

\subsection{Dataset Construction}
\subsubsection{Chinese Grid-style Corpus}
We follow the design of DB-MMLC \cite{lombard_2}. The Chinese corpus in MLD-VC consists of GRID-style Mandarin sentences composed of five components in a fixed order, namely name, verb, classifier, adjective, and noun. For each component, we provide 20 phonemically balanced candidate words and construct sentences by randomly sampling from these candidates. Tab. \ref{chinese-cropus} lists all candidate words. Note that the randomly generated sentences do not carry any actual meaning.

\begin{CJK}{UTF8}{gbsn}
\begin{table}[t]
\centering
\caption{The candidate phoneme-balanced word and the corresponding Pinyin.}
\resizebox{\linewidth}{!}{\begin{tabular}{lllll}
\toprule[1pt]
\multicolumn{1}{c}{\textbf{Name}} &
\multicolumn{1}{c}{\textbf{Verb}} &
\multicolumn{1}{c}{\textbf{Classifier}} &
\multicolumn{1}{c}{\textbf{Adjective}} &
\multicolumn{1}{c}{\textbf{Noun}} \\
\toprule[1pt]
旭峰 (Xufeng) & 买 (Mai) & 零个 (Lingge) & 大 (Da) & 沙发 (Shafa) \\
青木 (Qingmu) & 乘 (Cheng) & 一架 (Yijia) & 小 (Xiao) & 飞机 (Feiji) \\
林俊 (Linjun) & 坐 (Zuo) & 两只 (Liangzhi) & 旧 (Jiu) & 火车 (Huoche) \\
建树 (Jianshu) & 去 (Qu) & 三条 (Santiao) & 快 (Man) & 菠萝 (Boluo) \\
郭浩 (Guohao) & 拿 (Na) & 四段 (Siduan) & 慢 (Man) & 枕头 (Zhentou) \\
南月 (Nanyue) & 养 (Yang) & 五棵 (Wuke) & 新 (Xin) & 床 (Chuang) \\
赵坤 (Zhaokun) & 来 (Lai) & 六艘 (Liusou) & 硬 (Ying) & 村 (Cun) \\
文路 (Wenlu) & 给 (Gei) & 七斤 (Qijin) & 软 (Ruan) & 蝴蝶 (Hudie) \\
范畴 (Fanchou) & 挥 (Hui) & 八两 (Baliang) & 胖 (Pang) & 鹅 (E) \\
宋坏 (Songhuai) & 拔 (Ba) & 九碗 (Jiuwan) & 瘦 (Shou) & 鸭 (Ya) \\
孙西 (Sunxi) & 踢 (Ti) & 零杯 (Lingbei) & 长 (Chang) & 平板 (Pingban) \\
弘扬 (Hongyang) & 催 (Cui) & 一瓶 (Yiping) & 高 (Gao) & 水杯 (Shuibei) \\
启辰 (Qichen) & 蹲 (Dun) & 二听 (Erting) & 甜 (Tian) & 西瓜 (Xigua) \\
加号 (Jiahao) & 看 (Kan) & 三根 (Sangen) & 热 (Re) & 台灯 (Taideng) \\
壮硕 (Zhuangshuo) & 啃 (Ken) & 四张 (Sizhang) & 香 (Xiang) & 表格 (Biaoge) \\
明惠 (Minghui) & 抱 (Bao) & 五枚 (Wumei) & 乱 (Luan) & 电池 (Dianchi) \\
凌翔 (Lingxiang) & 喝 (He) & 六则 (Liuze) & 轻 (Qing) & 大米 (Dami) \\
云妮 (Yunni) & 试 (Shi) & 七顿 (Qidun) & 方 (Fang) & 键盘 (Jianpan) \\
佳倩 (Jiaqian) & 吃 (Chi) & 八匹 (Bapi) & 圆 (Yuan) & 面条 (Miantiao) \\
耕田 (Gengtian) & 种 (Zhong) & 九位 (Jiuwei) & 细 (Xi) & 字帖 (Zitie) \\
\bottomrule[1pt]
\end{tabular}
}
\label{chinese-cropus}
\end{table}
\end{CJK}

\subsubsection{Recoding Environment}
\begin{figure}[t]
  \centering
   \includegraphics[width=0.95\linewidth]{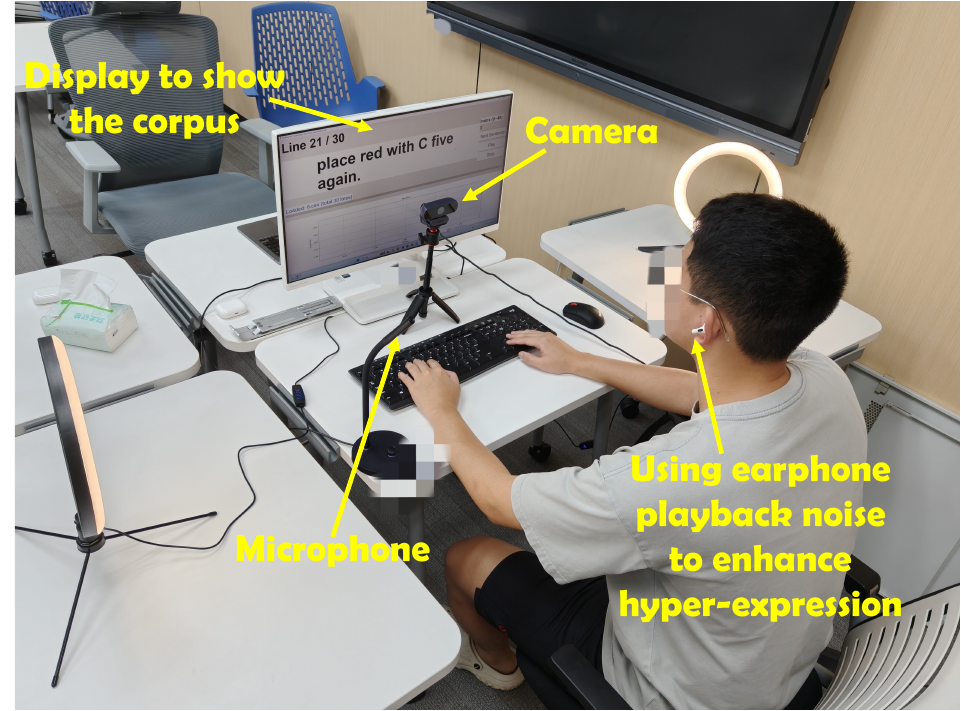}
   \caption{Picture of the recording environment.}
   \label{classroom}
\end{figure}

The MLD-VC dataset was recorded in a classroom located next to a corridor without any soundproofing. The building that houses the classroom is inside the campus, where nearby buildings were under construction and airplanes frequently flew overhead. As a result, the microphones unavoidably captured various types of environmental noise, including footsteps, conversations, passing cars, aircraft rumble, wind, and construction noise. To preserve the authenticity of the meeting scenario, we deliberately avoided applying any noise reduction techniques to suppress these sounds in the recorded speech. The recording sessions were conducted from 9 a.m. to 9 p.m. over several consecutive days. With this setup, the data in MLD-VC depict realistic video conference conditions. Fig.~\ref{classroom} illustrates the specific classroom setting used for recording.

\subsubsection{Subdataset Duration}
\begin{figure}[t]
  \centering
   \includegraphics[width=0.95\linewidth]{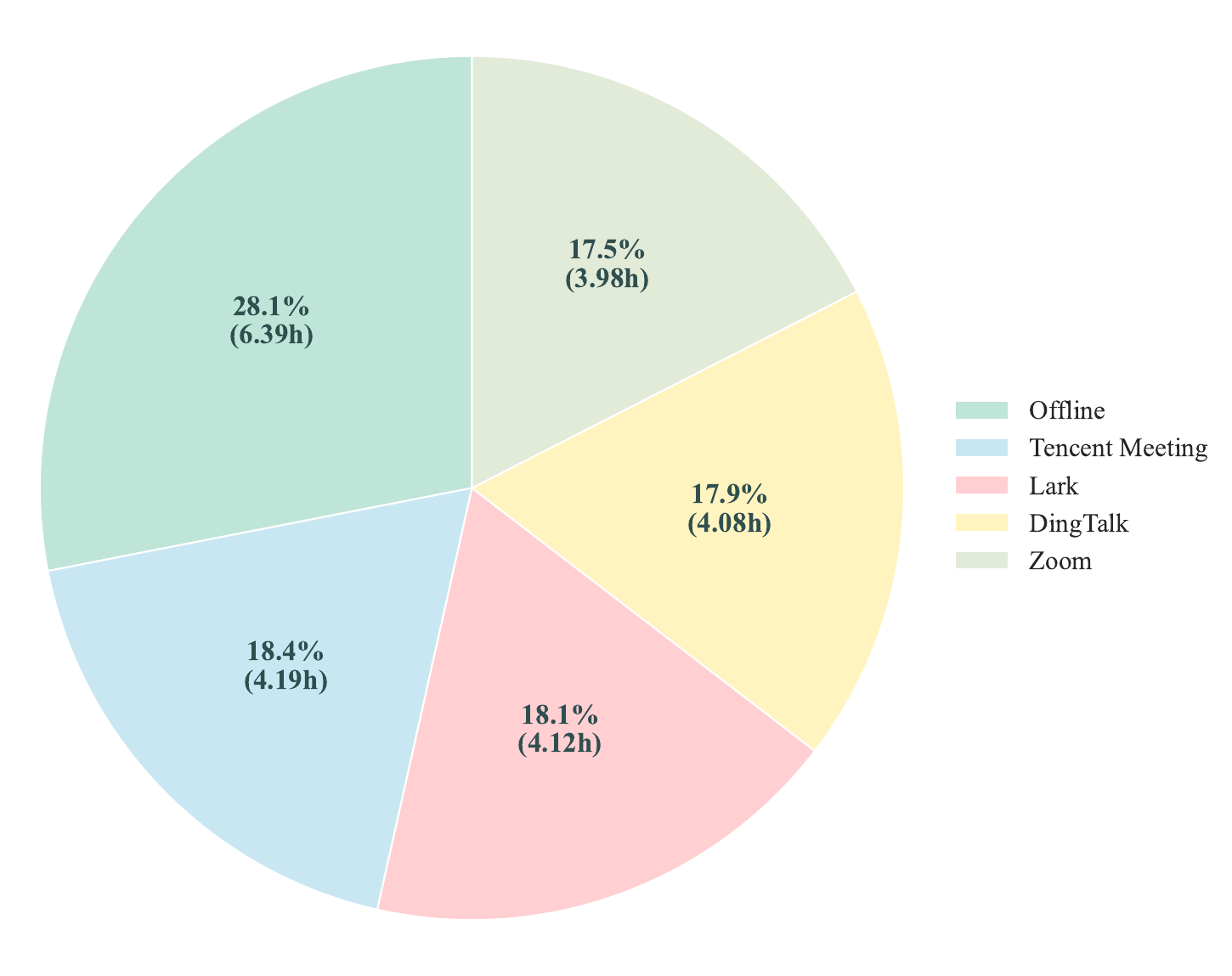}
   \caption{Duration distribution across subsets of the proposed MLD-VC dataset. “Offline” refers to the subset of recorded content that was captured before video conferencing.}
   \label{subdataset_duration}
\end{figure}
Fig.~\ref{subdataset_duration} presents the duration of each subset in the MLD-VC dataset. The “Offline” subset contains data that is not transmitted through any platform while still preserving the hyper-expression effect. Each of the remaining subsets corresponds to a specific video conferencing platform and therefore reflects both platform transmission and the presence of the hyper-expression effect.
\subsection{Visiual Feature Analysis}
\subsubsection{Metric}
\begin{figure}[t]
  \centering
   \includegraphics[width=0.95\linewidth]{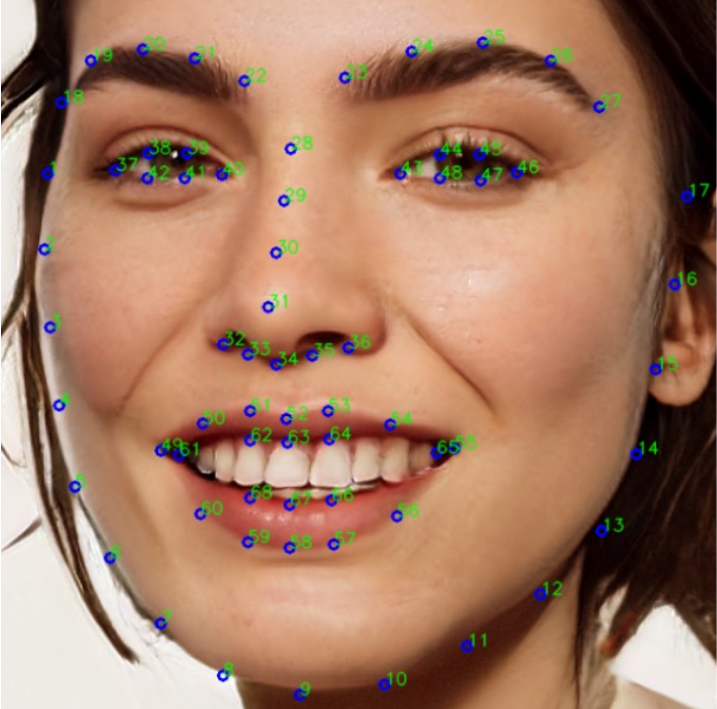}
   \caption{The illustration of face landmarks.}
   \label{face_landmark}
\end{figure}
For the analysis of visual features, we do not use traditional image quality metrics such as PSNR and SSIM. This is because compression and network latency in online transmission inevitably degrade image quality and can even cause blurring, so abnormal values of these metrics are unavoidable. For the visual input, the core objective of AVSR models is to recognize lip motion. Therefore, we select lip width, lip height, and lip roundness as visual metrics that are directly related to lip movement.

Each metric is computed from facial landmark locations. Fig.~\ref{face_landmark} shows the facial landmarks and their indices. In implementation, we first detect 68 landmarks on each video frame and normalize the scale of all landmark coordinates using the interocular distance. Specifically, we compute the average of the landmarks of the left eye (points 37 to 42) and the right eye (points 43 to 48) to obtain the centers of the two eyes, take their Euclidean distance as the interocular distance, and use this distance as a normalization factor to rescale all facial landmark coordinates. This removes scale variations caused by different speakers and camera distances.

\begin{figure}[t]
  \centering
   \includegraphics[width=1\linewidth]{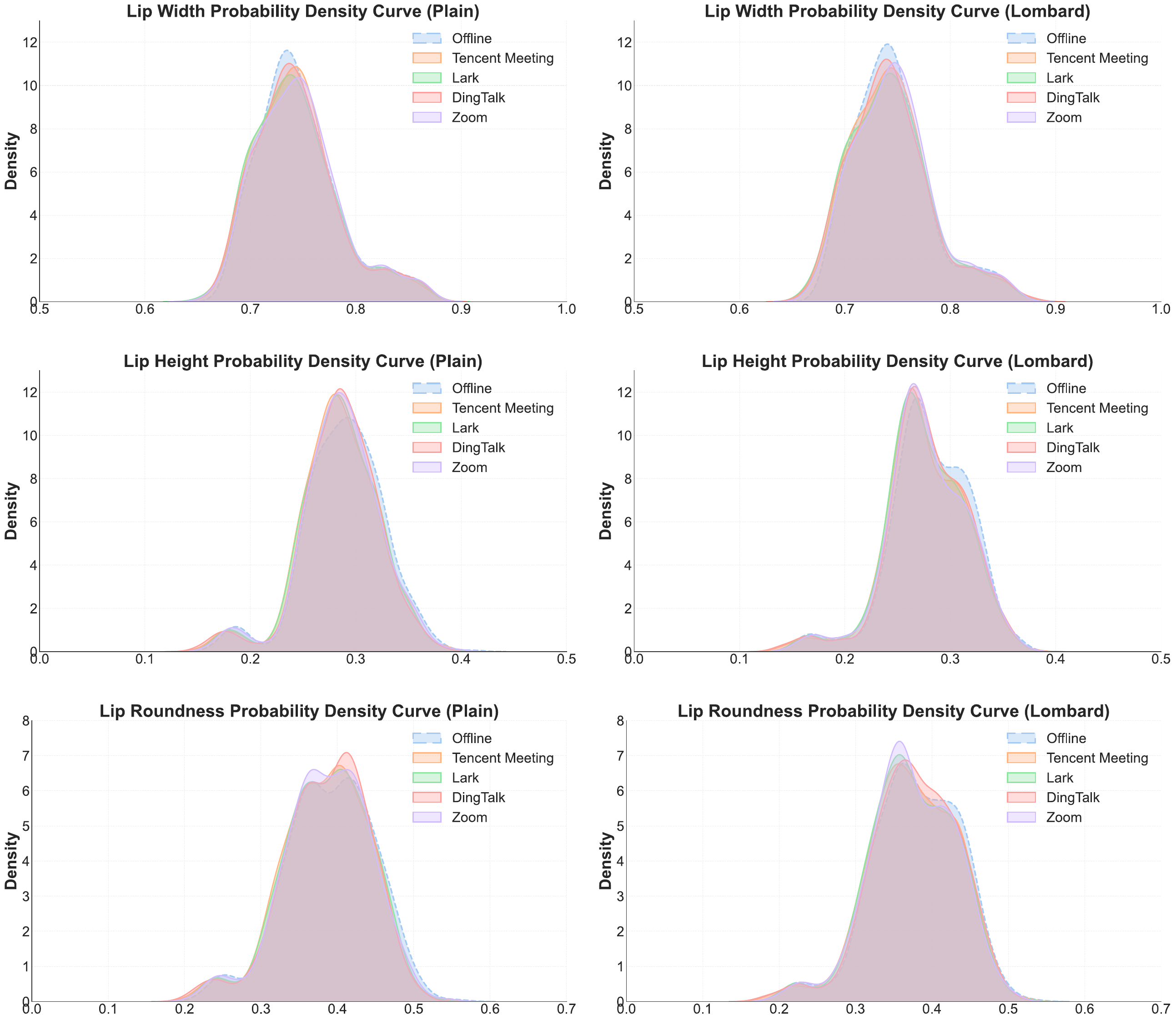}
\caption{
Probability density curves of three visual features (lip width, lip height, and lip roundness) across five subsets in the proposed MLD-VC under \textit{Plain} (left column) and \textit{Lombard} (right column).
}
\label{problem_location_vision}
\end{figure}

\begin{table}[t]
\centering
\caption{Peak abscissas of probability density curves for different visual features.}
\label{peak_x_coordinates_visual}
\resizebox{\linewidth}{!}{%
\begin{tabular}{cccccc}
\toprule
\multirow{2}{*}{\textbf{Visual Feature}} & \multicolumn{5}{c}{\textbf{Platform}} \\
\cmidrule(lr){2-6}
 & \textbf{Offline} & \textbf{Tencent Meeting} & \textbf{Lark} & \textbf{DingTalk} & \textbf{Zoom} \\
\midrule
Lip Width (\textit{clean})  & 0.73 & 0.74 & 0.74 & 0.74 & 0.75 \\
Lip Width (\textit{80 dB})  & 0.74 & 0.74 & 0.74 & 0.74 & 0.75 \\
\midrule
Lip Height (\textit{clean})  & 0.29 & 0.28 & 0.28 & 0.29 & 0.28 \\
Lip Height (\textit{80 dB})  & 0.27 & 0.26 & 0.26 & 0.27 & 0.27 \\
\midrule
Lip Roundness (\textit{clean})  & 0.41 & 0.40 & 0.40 & 0.41 & 0.37 \\
Lip Roundness  (\textit{80 dB})  & 0.36 & 0.36 & 0.36 & 0.36 & 0.36 \\
\bottomrule
\end{tabular}%
}
\end{table}

\begin{figure*}[t]
  \centering
   \includegraphics[width=0.8\linewidth]{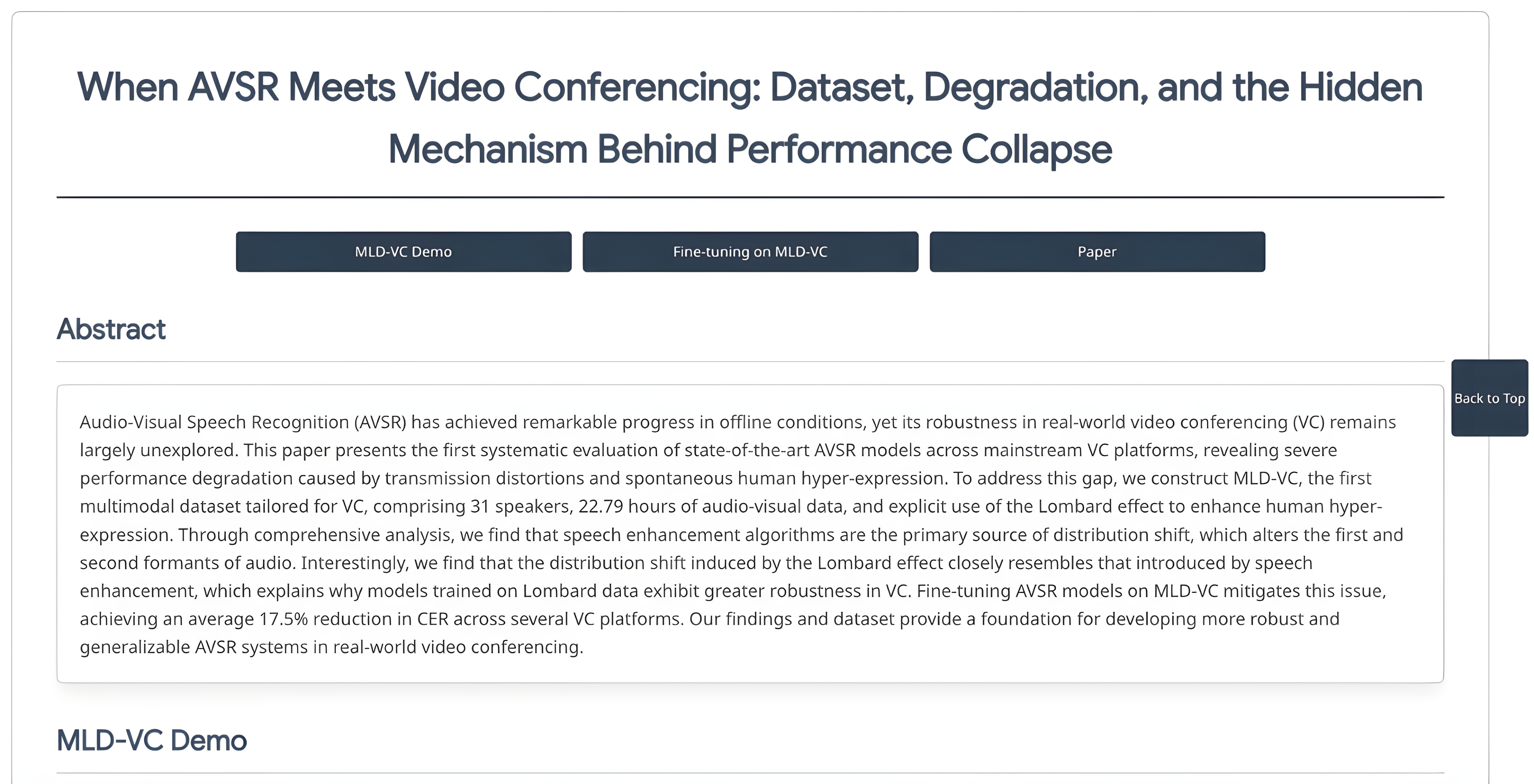}
\caption{The static page MLD-VC demo in the supplementary material.}
\label{page_demo}
\end{figure*}
In the normalized coordinate system, we compute lip width, lip height, and lip roundness from the lip landmarks. First, we define lip width $W$ as the Euclidean distance between the left and right mouth corners (points 49 and 55). Second, we compute the average vertical coordinates of the upper lip point set \{51, 52, 53, 62, 63, 64\} and the lower lip point set \{57, 58, 59, 66, 67, 68\}, and define lip height $H$ as the absolute difference between these two averages. On this basis, we define lip roundness $C$ as the ratio between height and width, that is $C = \frac{H}{W}$. This metric reflects the relative proportion of the lip shape in the vertical and horizontal directions, where values of $C$ closer to 1 indicate that the lip shape is closer to a circle.

\subsubsection{Results}

We plot the probability density curves of the visual features in Fig. \ref{problem_location_vision}. In addition, Tab. \ref{peak_x_coordinates_visual} lists the horizontal coordinate of the peak for each curve. The results show that the three selected visual features exhibit no obvious change between the offline condition and the various video conferencing platforms. This indicates that landmark-based visual geometric features are robust in video conferencing scenarios. Therefore, we recommend using these stable visual features as the input to the visual stream rather than relying solely on unstable lip images in the video conferencing scenarios.

\subsection{Implement Details of Fine-tuning}
To ensure a fair comparison, we maintained consistent hyperparameters between the fine-tuning and ablation experiments. The learning rate was set to 0.0001, the number of fine-tuning steps was set to 800, the weight decay was set to 0.01, and the dropout rate was set to 0.3. Besides, to prevent overfitting in the fine-tuned model, we added data from the original training set. Specifically, the ratio between fine-tuning data and original training data was 7 to 3.

\section{Dataset Demo}

We built a local static web page to present demos of our dataset and included it in the compressed supplementary material package, namely "supplementary-material.zip". In this web demo, we present recordings from two speakers of different genders, captured on different video conferencing platforms, in different languages, and under different noise conditions to enhance hyper-expression. Fig.~\ref{page_demo} illustrates the static web page. We recommend that you view the file named “demo.html” in the supplementary material to gain a better understanding of MLD-VC.

\end{document}